\documentclass{article}

     \usepackage[nonatbib, preprint]{neurips_2020}

\usepackage[utf8]{inputenc} 
\usepackage[T1]{fontenc}    
\usepackage{hyperref}       
\usepackage{url}            
\usepackage{booktabs}       
\usepackage{amsfonts}       
\usepackage{nicefrac}       
\usepackage{microtype}      
\usepackage[english]{babel}
\usepackage{graphicx}
\usepackage{xcolor}
\usepackage[backend=biber,
style=numeric, maxbibnames=99, url=false]{biblatex}

\hypersetup{
    colorlinks=true,
    linkcolor=blue,
    filecolor=magenta,      
    urlcolor=cyan,
}

\title{Predicting Video with VQVAE}

\author{
   Jacob Walker \\
   DeepMind \\
   \texttt{jcwalker@google.com} \\
   \And
   Ali Razavi \\
   DeepMind \\
   \texttt{alirazavi@google.com} \\
   \And
   Aäron van den Oord \\
   DeepMind \\
   \texttt{avdnoord@google.com} \\
}

\begin{document}

\maketitle

\begin{abstract}
  In recent years, the task of video prediction---forecasting future video given past video frames---has attracted attention in the research community. In this paper we propose a novel approach to this problem with Vector Quantized Variational AutoEncoders (VQ-VAE). With VQ-VAE we compress high-resolution videos into a hierarchical set of multi-scale discrete latent variables. Compared to pixels, this compressed latent space has dramatically reduced dimensionality, allowing us to apply scalable autoregressive generative models to predict video. In contrast to previous work that has largely emphasized highly constrained datasets, we focus on very diverse, large-scale datasets such as Kinetics-600. We predict video at a higher resolution on unconstrained videos, $256\times256$, than any other previous method to our knowledge. We further validate our approach against prior work via a crowdsourced human evaluation.

\end{abstract}

\section{Introduction}
When it comes to real-world image data, deep generative models have made substantial progress. With advances in computational efficiency and improvements in architectures, it is now feasible to generate high resolution, realistic images from vast and highly diverse datasets \cite{BrockDS19, RazaviOV19, karras2017progressive}. Apart from the domain of images, deep generative models have also shown promise in other data domains such as music \cite{DielemanOS18, jukebox}, speech synthesis \cite{oord2016wavenet}, 3D voxels \cite{LiuGO18, NashW17}, and text \cite{radford2019language}. One particular fledgling domain is video.

While some work in the area of video generation \cite{clark2020adversarial, VondrickPT16, saito2018} has explored video synthesis---generating videos with no prior frame information---many approaches actually focus on the task of video prediction conditioned on past frames \cite{RanzatoSBMCC14, SrivastavaMS15, patraucean2015spatiotemporal, MathieuCL15, lee2018savp, babaeizadeh2018stochastic, OliuSE18, XiongL00L18, Xue0BF16, Finn16, luc2020}. It can be argued that video synthesis is a combination of image generation and video prediction. In other words, one could decouple the problem of video synthesis into unconditional image generation and conditional video prediction from a generated image. Therefore, we specifically focus on video prediction in this paper. Potential computer vision applications of video forecasting include interpolation, anomaly detection, and activity understanding. More generally, video prediction also has more general implications for intelligent systems---the ability to anticipate the dynamics of the environment. The problem is thus also relevant for robotics and reinforcement learning \cite{Finn16, Ebert17, OhGLLS15, ha2018world, RacaniereWRBGRB17}.

Approaches toward video prediction have largely skewed toward variations of generative adversarial networks \cite{MathieuCL15, lee2018savp, clark2020adversarial, VondrickPT16, luc2020}. In comparison, we are aware of only a relatively small number of approaches which propose variational autoencoders \cite{babaeizadeh2018stochastic, Xue0BF16, denton18a}, autoregressive models \cite{KalchbrennerOSD17, WeissenbornTU20}, or flow based approaches \cite{KumarBEFLDK20}. There may be a number of reasons for this situation. One is the explosion in the dimensionality of the input space. A generative model of video needs to model not only one image but tens of them in a coherent fashion. This makes it difficult to scale up such models to large datasets or high resolutions. In addition, previous work \cite{clark2020adversarial} suggests that video prediction may be fundamentally more difficult than video synthesis; a synthesis model can generate simple samples from the dataset while prediction potentially forces the model to forecast conditioned on videos that are outliers in the distribution. Furthermore, most prior work has focused on datasets with low scene diversity such as Moving MNIST \cite{SrivastavaMS15}, KTH \cite{KTH}, or robotic arm datasets \cite{Finn16, Ebert17}.
While there have been attempts to \emph{synthesize} video at a high resolution \cite{clark2020adversarial}, we know of no attempt---excluding flow based approaches---to \emph{predict} unconstrained video beyond resolutions of 64x64.

In this paper we address the large dimensionality of video data through compression. Using Vector Quantized Variational Autoencoders (VQ-VAE) \cite{OordVK17}, we can compress video into a space requiring only 1.3\% of the bits expressed in pixels. While this compressed encoding is lossy, we can still reconstruct the original video from the latent representation with a high degree of fidelity. Furthermore, we can leverage the modularity of VQ-VAE and decompose our latent representation into a hierarchy of encodings, separating high-level, global information from details such as fine texture or small motions. Instead of training a generative model directly on pixel space, we can instead model this much more tractable discrete representation, allowing us to train much more powerful models, use large diverse datasets, and generate at a high resolution. While most prior work has focused on GANs, this discrete representation can also be modeled by likelihood-based models. Likelihood models in concept do not suffer from mode-collapse, instability in training, and lack of diversity of samples often witnessed in GANs \cite{denton18a, babaeizadeh2018stochastic, RazaviOV19}. In this paper, we propose a PixelCNN augmented with causal convolutions in time and spatiotemporal self-attention to model this space of latents. In addition, because the latent representation is decomposed into a hierarchy, we can exploit this decomposition and train separate specialized models at different levels of the hierarchy. 

Our paper makes four contributions. First, we demonstrate the novel application of VQ-VAE to video data. Second, we propose a set of spatiotemporal PixelCNNs to predict video by utilizing the latent representation learned with VQ-VAE. Third, we explicitly predict video at a higher resolution on than ever before on real-world, unconstrained video. Finally, we demonstrate the competitive performance of our model with a crowdsourced human evaluation.

\begin{figure}
  \begin{tabular}{c||cc}
  \includegraphics[width=40mm]{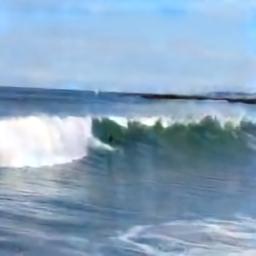} &
  \includegraphics[width=40mm]{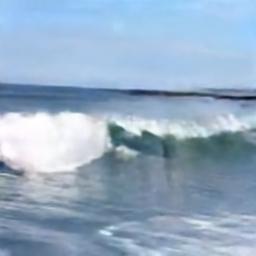} &
  \includegraphics[width=40mm]{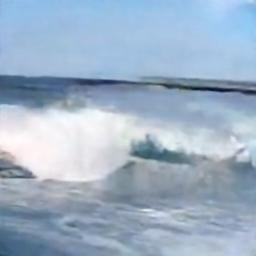} \\ 
  4th Frame & 9th Frame & 16th Frame \\
  \end{tabular}
  \caption{In this paper we predict video at a high resolution ($256\times256$) using a compressed latent representation. The first 4 frames are given as conditioning. We predict the next 12, two of which (9th and 16th) we show on the right. All frames shown here have been compressed by VQ-VAE. Videos licensed under CC-BY. Attribution for videos in this paper can be found in section~\ref{Attributions}. Best seen in video on our website \href{https://sites.google.com/view/predicting-video-with-vqvae/home}{here}.}
\label{teaser}
\end{figure}

\section{Background}
\begin{figure}
  \centering
  \begin{tabular}{ccc}
  \includegraphics[width=40mm]{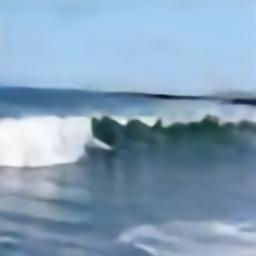} &
  \includegraphics[width=40mm]{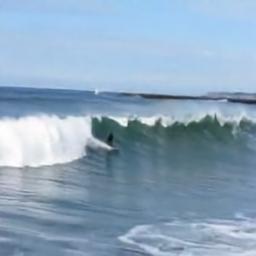} &
  \includegraphics[width=40mm]{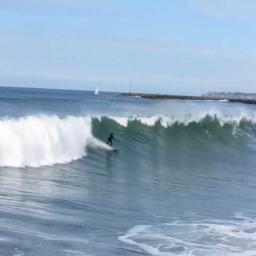} \\   
  Top Layer & Top + Bottom Layer & Original Frame \\
  \end{tabular}
  \caption{Here we demonstrate the compression capability of VQ-VAE. The top and bottom rows represent two different frames within the same video. The top layer retains most of the global information, while the bottom layer adds fine detail. Videos licensed under CC-BY. Attribution for videos in this paper can be found in section~\ref{Attributions}.}
\label{reconst}
\end{figure}

\subsection{Vector Quantized Autoencoders}
VQ-VAEs \cite{OordVK17} are autoencoders which learn a discrete latent encoding for input data $x$. First, the output of non-linear encoder $z_{e}(x)$, implemented by a neural network, is passed through a discretization bottleneck. $z_{e}(x)$ is mapped via nearest-neighbor into a quantized codebook $e \in R^{K \times D}$ where $D$ is the dimensionality of each vector $e_{j}$ and $K$ is the number of categories in the codebook. The discretized representation is thus given by:

\begin{equation}
    z_{q}(x) =  e_{k} \; \textrm{where} \; k = \textrm{argmin}_{j}||z_{e}(x) - e_{j}||_{2}
\label{vq1}
\end{equation}

Equation \ref{vq1} is not differentiable; however, \cite{OordVK17} notes that copying the gradient of $z_{q}(x)$ to $z_{e}(x)$ is a suitable approximation similar to the straight-through estimator \cite{BengioLC13}. A decoder $D$, also implemented by a neural network, then reconstructs the input from $z_{q}(x)$. The total loss function for the VQ-VAE is thus:

\begin{equation}
L = ||D(z_{g}(x)) - x||_{2}^{2} + ||\textrm{sg}[z_{g}(x)] - \mathbf{e}||_{2}^{2} + \beta||z_{g}(x) - \textrm{sg}[\mathbf{e}]||_{2}^{2}
\label{vq2}
\end{equation}

Where sg is a stop gradient operator, and $\beta$ is a parameter which regulates the rate of code change. As in previous work \cite{OordVK17, RazaviOV19}, we replace the second term in equation \ref{vq2} and learn the codebook $e \in R^{K \times D}$ via an exponential moving average of previous values during training:

\begin{equation}
N_{i}^{(t)} := N_{i}^{(t-1)}*\gamma + n_{i}^{(t)}(1-\gamma),  \; m_{i}^{(t)} := m_{i}^{(t-1)}*\gamma+\sum_{j}^{n_{i}^{(t)}}z_{e}(x)_{i,j}^{(t)}(1-\gamma), \; e_{i}^{(t)} := \frac{m_{i}^{(t)}}{N_{i}^{(t)}}
\label{vq3}
\end{equation}

Where $\gamma$ is a decay parameter and $n_{i}^{(t)}$ is the numbers of vectors in $z_{g}(x)$ in a batch that will map to $e_{i}$.

\subsection{PixelCNN Models}

PixelCNN and related models have shown promise in modeling a wide variety of data domains \cite{OordKK16, oord2016wavenet, KalchbrennerOSD17, WeissenbornTU20}. These autoregressive models are likelihood-based---they explicitly optimize negative log-likelihood. They exploit the fact that the joint probability distribution input data $x$ can be factored into a product of conditional distributions for each dimension of the data:

\begin{equation}
P_{\theta}(\mathbf{x}) = \prod_{i=0}^{n}p_{\theta}(x_{i}|x_{<i})
\label{pcnn}
\end{equation}

Where $n$ is the full dimensionality of the data. This factorization is implemented by a neural network, and the exact set of conditional dependencies is determined by the data domain. Image pixels may depend on regions above and to the left of them \cite{OordKK16}, while temporal dimensions may depend on past dimensions \cite{oord2016wavenet, KalchbrennerOSD17, WeissenbornTU20}.

\section{Method}

\begin{figure}
\begin{tabular}{c||c}
  \includegraphics[width=65mm, height=45mm, trim=38mm 60mm 55mm 20mm, clip]{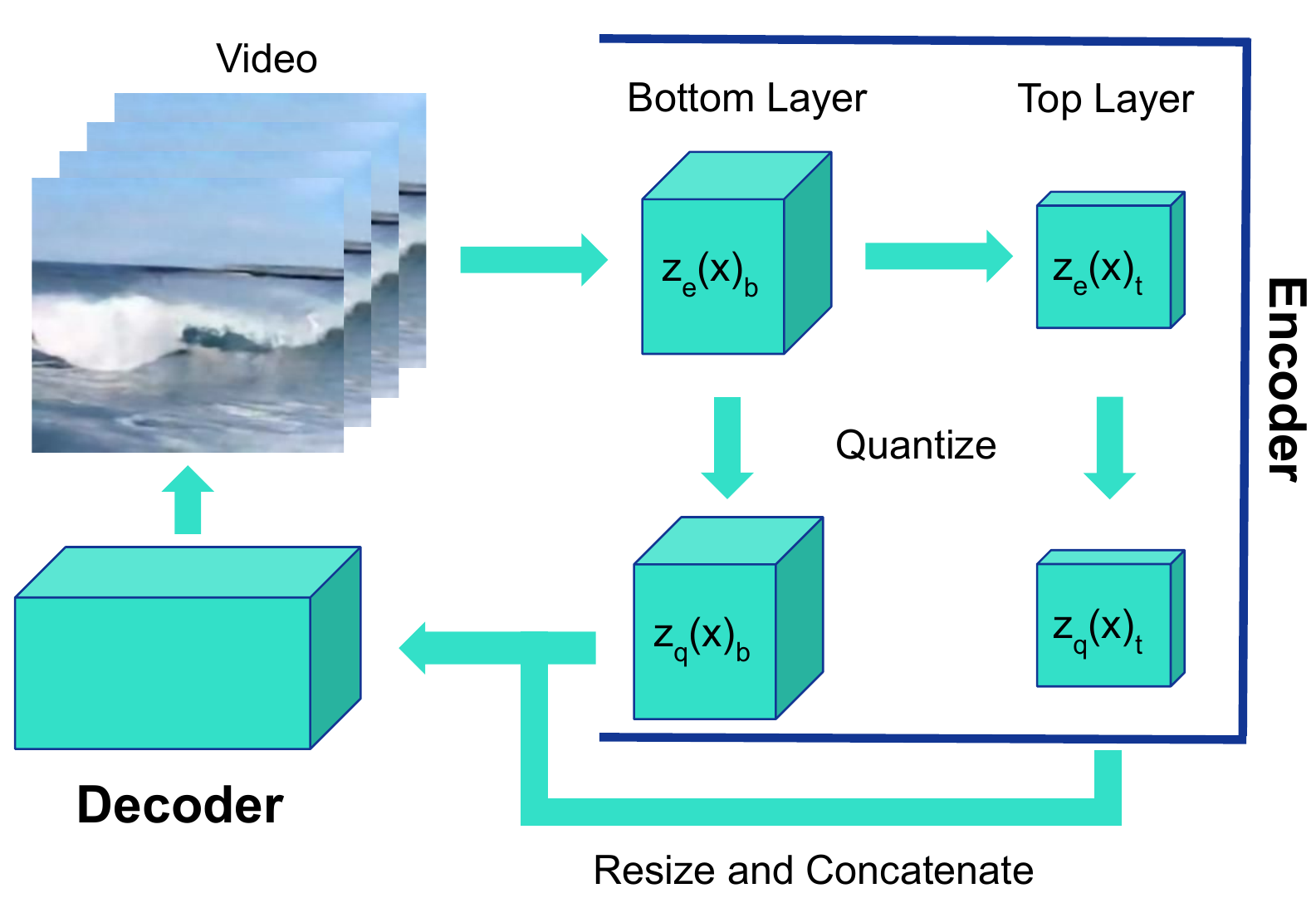} &
  \includegraphics[width=65mm, height=45mm, trim=20mm 65mm 60mm 10mm, clip]{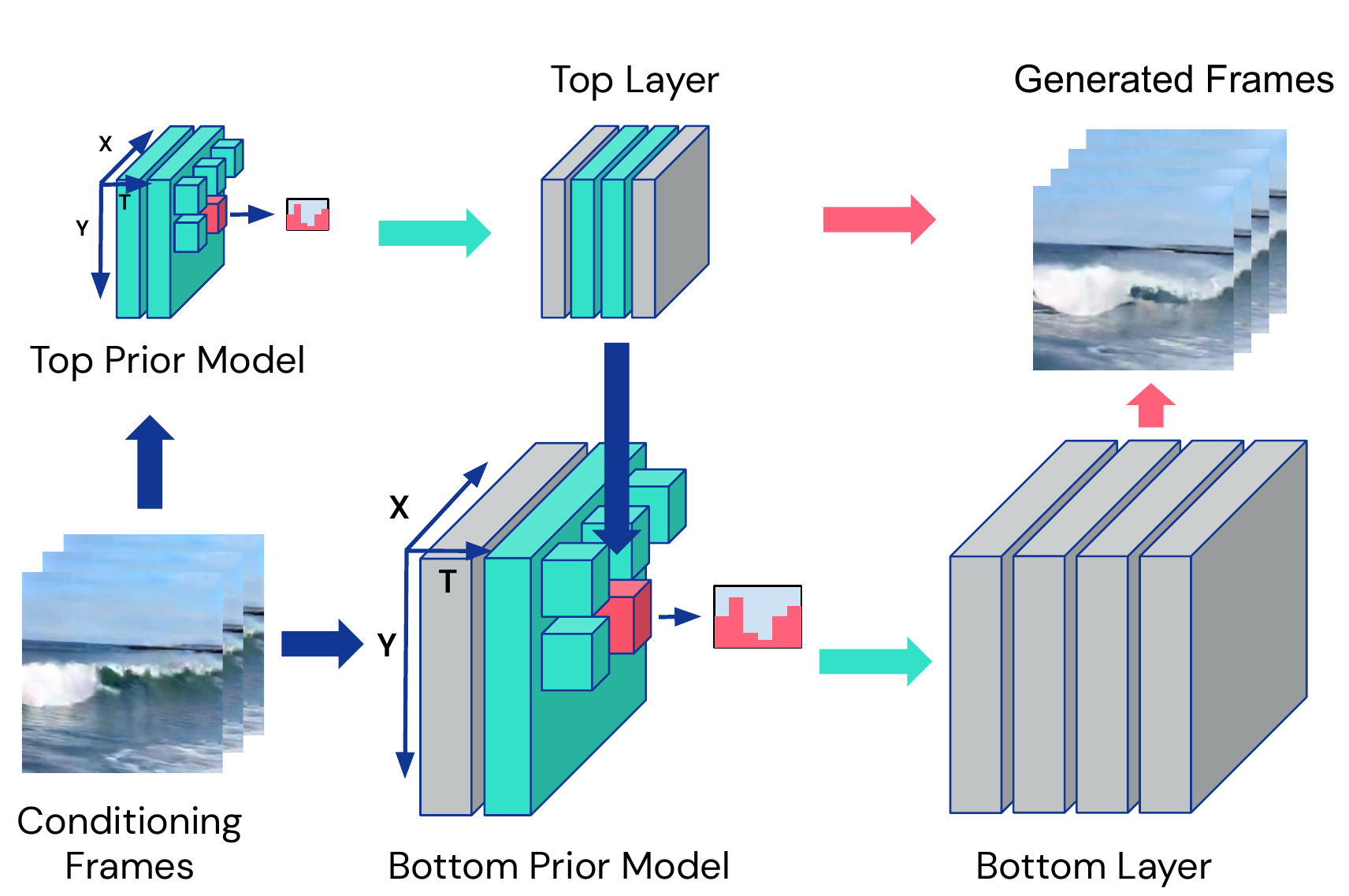} \\
  VQ-VAE Model & Predicting Video
  \end{tabular}
  \caption{Here we show an overview of our approach. On the left we show the process of compressing video with VQ-VAE. On the right we show the process of generating video with the latents. The top conditional prior model is a PixelCNN with causal convolutions to incorporate all past information at each point in space-time. The bottom conditional prior model is simply a 2D PixelCNN which generates slice by slice. It is conditioned with a convolutional tower which incorporates a window of time slices from the top latents and past bottom latents. The slices outside of this window are colored grey in this diagram. Blue arrows represent conditioning, green arrows generation, and pink feed-forward decoding. Videos licensed under CC-BY. Attribution for videos in this paper can be found in section~\ref{Attributions}.}
\label{VQVAE}
\end{figure}

Our approach consists of two main components. First, we compress video segments into a discrete latent representation using a hierarchical VQ-VAE. We then propose a multi-stage autoregressive model based on the PixelCNN architecture, exploiting the low dimensionality of the compressed latent space and the hierarchy of the representation. 

\subsection{Compressing Video with VQ-VAE}
Similar to \cite{RazaviOV19}, we use VQ-VAE to compress video in a hierarchical fashion. This multi-stage composition of the latent representation allows decomposition of global, high level information from low-level details such as edges or fine motion. For image information \cite{RazaviOV19} this approach confers a number of advantages. First, the decomposition allows latent codes to specialize at each level. High level information can be represented in an even more compressed manner, and the total reconstruction error is lower. In addition, this hierarchy leads to a naturally modular generative model. We can then develop a generative model that specializes in modeling the high-level, global information. We can then train a separate model, conditioned on global information, that fills in the details and models the low-level information further down the hierarchy. In this paper, we adopt the terminology of \cite{RazaviOV19} and call the set of high-level latents the \emph{top} layer and the low-level latents the \emph{bottom} layer. 

Consistent with the experimental setup of previous work in video prediction, we deal with 16-frame videos. Most of the videos in our training dataset are 25 frames per second. We use frames at a $256\times256$ resolution. The full video voxel is thus $256\times256\times16$. Using residual blocks with 3D convolutions, we downsample the video spatiotemporally. At the bottom layer, the video is downsampled to a quantized latent space of $64\times64\times8$, reducing the spatial dimension by 4 and the temporal dimension by 2. Another stack of blocks reduces all dimensions by 2, with a top layer of $32\times32\times4$. Each of the voxels in the layer is quantized into 512 codes with a different codebook for both layers.

The decoder then concatenates the bottom layer and the top layer after upsampling using transposed convolutions. From this concatentation as input, the decoder deterministically outputs the full $256\times256\times16$ video. Overall, we reduce a $256\times256\times16\times3\times\log(256)$ space down to a $64\times64\times8\times\log(512) + 32\times32\times4\times\log(512)$ space, a greater than 98\% reduction in bits required.  During training, we randomly mask out the bottom layer in the concatenated input to the decoder. Masking encourages the model to utilize the top latent layer and prevent codebook collapse.

\subsection{Predicting Video with PixelCNNs}
With VQ-VAE, our $256\times256\times16$ video is now decomposed into a hierarchy of quantized latents at $64\times64\times8$ and $32\times32\times4$. Previous autoregressive approaches involved full pixel videos at $64\times64\times16$ \cite{WeissenbornTU20} or  $64\times64\times20$ \cite{KalchbrennerOSD17}. Our latent representation is thus well within the range of tractability for these models. Furthermore, given the hierarchy, we can factorize our generative model into a coarse-to-fine fashion. We denote the model of the top layer the \emph{top prior} and model of the bottom layer the \emph{bottom prior}. Because we are focusing on video prediction, we emphasize that both are still conditioned on a series of input frames. While previous work used 5 frames and predicted 11 \cite{WeissenbornTU20, clark2020adversarial}, the power-of-two design of our architecture leads us to condition on 4 and predict 12. When conditioning our prior models on these frames, we need not use a large stack directly on the original images but save memory and computation by training a smaller stack of residual layers on their latent representation, compressing these 4 conditional frames into a small latent space of $32\times32$ and $64\times64\times2$.

We first model the top layer with a conditional prior model. Our prior model is based on a PixelCNN with multi-head self attention layers \cite{ChenMRA18}. We adapt this architecture by extending the PixelCNN into time; instead of a convolutional stack over a square image, we use a comparable 3D convolutional stack over the cube representing the prior latents. The convolutions are masked in the same way as the original PixelCNN in space---at each location in space, the convolutions only have access to information to the left and above them. In time, present and future timesteps are masked out, and convolutions only have access to previous timesteps. While spatiotemporal convolutions can be resource intensive, we can implement most of this functionality with 2D convolutions separately in the $x-t$ plane and the $y-t$ plane. We take 1D convolutions in the horizontal and vertical stacks in the original PixelCNN \cite{OordKK16} and add the extra dimension of time to them, making them 2D. Our only true 3D convolution is at the first layer before the addition of gated stacks. We use multi-head attention layers analogous to \cite{RazaviOV19}; this time the attention is applied to a 3D voxel instead of a 2D layer as in \cite{RazaviOV19}. Attention is applied every five layers. During sampling we can generate voxels left-to-right, top-to-bottom within each temporal step as in the original PixelCNN. Once a final step is generated, we can generate the next step conditioned on the previous generated steps. 

Once we have a set of latents from the top layer, we can condition our bottom conditional prior model and generate the final bottom layer. Because the bottom layer has a higher number of dimensions and relies on local information, we don't necessarily need a 3D PixelCNN. Instead, we use a 2D PixelCNN with multi-head self attention every five layers analogous to \cite{RazaviOV19}. We implement a 3D conditional stack, however, that takes in a window of time steps from the top layer as well as a window of past generated time steps in the bottom layer. The window sizes we used were 4 and 2 respectively. This conditional stack is used as conditioning to the 2D PixelCNN at the current timestep. 
\section{Related Work}
\begin{table}
  \caption{Human Evaluation on Kinetics-600}
  \centering
  \begin{tabular}{ccc}
    \toprule
    Prefer Video VQ-VAE & Prefer \cite{luc2020} & Indifferent  \\
    \midrule
    65.7\% & 12.8\% & 21.5\% \\
    \bottomrule
  \end{tabular}
  \label{quant_results2}
\end{table}

\paragraph{Video Prediction and Synthesis:}
In the last few years, the research community has focused a spotlight on the topic of video generation---either in the form of video synthesis or prediction. Early approaches involved direct, deterministic pixel prediction \cite{RanzatoSBMCC14, SrivastavaMS15, OhGLLS15, patraucean2015spatiotemporal}. Given the temporal nature of video, such approaches often incorporated LSTMs. These papers usually applied their deterministic models on datasets such as moving MNIST characters \cite{SrivastavaMS15}; because of their deterministic nature, rarely were they successfully applied to more complex datasets. Given this situation, researchers started to adapt popular models for image generation to the problem starting with generative adversarial models \cite{MathieuCL15, VondrickPT16, lee2018savp, babaeizadeh2018stochastic, clark2020adversarial, saito2018, luc2020, XiongL00L18}, variational autoencoders \cite{Xue0BF16}, and autoregressive models \cite{KalchbrennerOSD17, WeissenbornTU20}. Others stepped aside from the problem of full pixel prediction and instead predicted pixel motion \cite{Finn16, Walker16, JiaBTG16} or a decomposition of pixels and motion \cite{denton18a, Gao_2019_ICCV, pmlr-v80-jang18a, hao2018controllable, LiFYWLY18, Tulyakov0YK18, VillegasICLR17}. Finally, some have proposed a hierarchical approach based on structured information---generating video conditioned on text \cite{LiMSCC18}, semantic segments \cite{LucNCVL17, LucCLV18}, or human pose \cite{Walker17, VillegasICML17}.

\paragraph{Compressing Data with Latents:}
The key element in our video prediction framework is compression---representing videos through lower dimensional latents. We apply the framework of VQ-VAE \cite{OordVK17, RazaviOV19} which has been successfully applied to compress image and sound data.
Related to VQ-VAE, other researchers have explored hierarchies of latents for generation of images \cite{defauw19} and music \cite{DielemanOS18}.

\paragraph{Autoregressive Models:}
The foundation of our model is based on PixelCNN \cite{OordKK16}. Distinct from implicit likelihood models such as GANs and approximate methods such as VAEs, the family of PixelCNN architectures have shown promise in modeling a variety of data domains including images \cite{OordKK16}, sound \cite{oord2016wavenet}, and video \cite{KalchbrennerOSD17, WeissenbornTU20}. In line with our paper, recent work with these models has shifted toward decomposing autoregression through hierarchies \cite{MenickK19, ReedOKCWCBF17} and latent compression \cite{OordVK17, RazaviOV19, jukebox}.

\section{Experiments}

\begin{table}
  \caption{FVD Scores on Kinetics-600. Lower is better.}
  \centering
  \begin{tabular}{ll}
    \toprule
    Method     & FVD Score ($\downarrow$) \\
    \midrule
    Video Transformer (64 $\times$ 64) \cite{WeissenbornTU20} & 170 $\pm$ 5	    \\
    DVD-GAN-FP (64 $\times$ 64) \cite{clark2020adversarial}   & 69.15 $\pm$ 1.16 \\
    TRIVD-GAN-FP (64 $\times$ 64) \cite{luc2020} & 25.74 $\pm$ 0.66  \\
    \midrule
    Video VQ-VAE (64 $\times$ 64) & 64.30 $\pm$ 2.04  \\
    Video VQ-VAE FVD* (64 $\times$ 64) & 54.30 $\pm$ 3.49  \\
    Video VQ-VAE (256 $\times$ 256) & 129.85 $\pm$ 1.64  \\
    Video VQ-VAE FVD* (256 $\times$ 256) & 82.45 $\pm$  1.16  \\
    \bottomrule
  \end{tabular}
  \label{quant_results}
\end{table}

\begin{figure}
  \begin{tabular}{c||cc}
  \includegraphics[width=40mm]{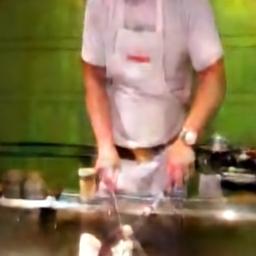} &
  \includegraphics[width=40mm]{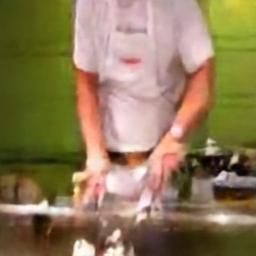} &
  \includegraphics[width=40mm]{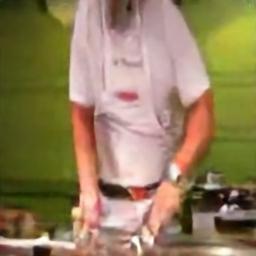} \\ \includegraphics[width=40mm]{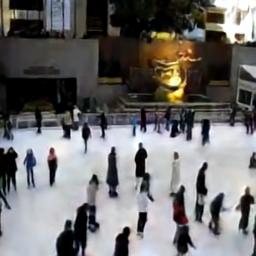} &
  \includegraphics[width=40mm]{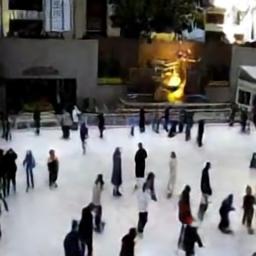} &
  \includegraphics[width=40mm]{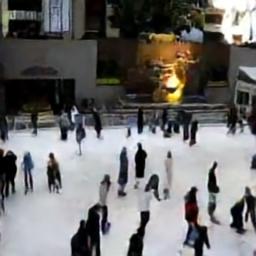} \\
  \includegraphics[width=40mm]{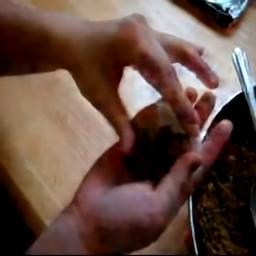} &
  \includegraphics[width=40mm]{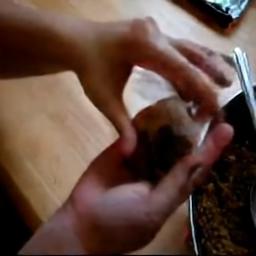} &
  \includegraphics[width=40mm]{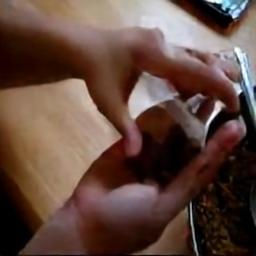} \\
  \includegraphics[width=40mm]{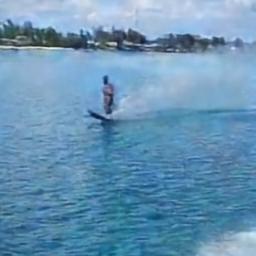} &
  \includegraphics[width=40mm]{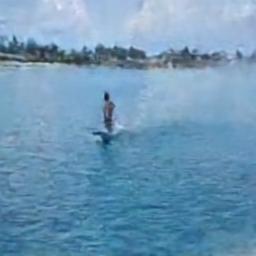} &
  \includegraphics[width=40mm]{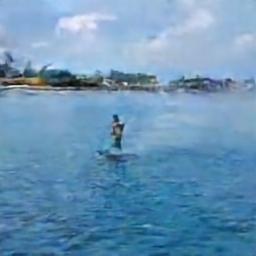} \\
  \includegraphics[width=40mm]{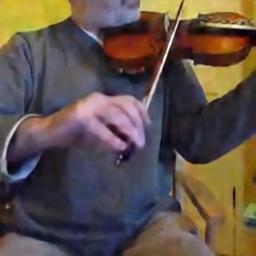} &
  \includegraphics[width=40mm]{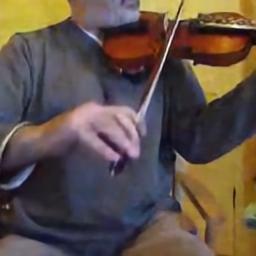} &
  \includegraphics[width=40mm]{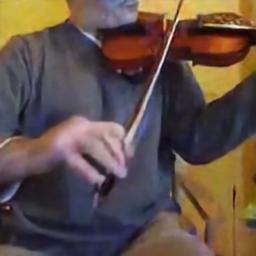} \\
  Final Conditioning Frame & 9th Frame & 16th Frame \\
  \end{tabular}
  \caption{Selected prediction results. The first 4 frames are given as conditioning. We predict the next 12, two of which (9th and 16th) we show on the right. All frames shown here have been compressed by VQ-VAE. Videos licensed under CC-BY. Attribution for videos in this paper can be found in section~\ref{Attributions}. Best seen in video on our website \href{https://sites.google.com/view/predicting-video-with-vqvae/home}{here}.}
\label{results1}
\end{figure}

\begin{figure}
  \begin{tabular}{c||cc}
  \includegraphics[width=40mm]{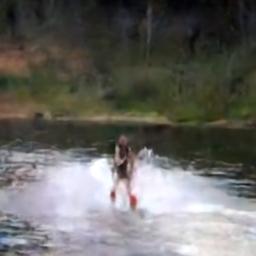} &
  \includegraphics[width=40mm]{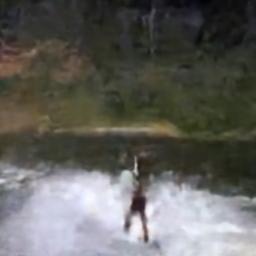} &
  \includegraphics[width=40mm]{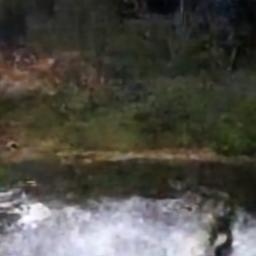} \\
  \includegraphics[width=40mm]{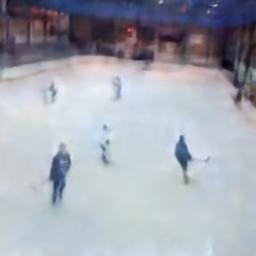} &
  \includegraphics[width=40mm]{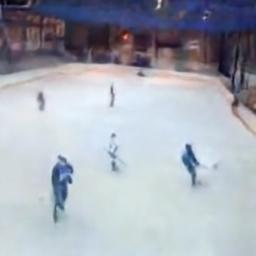} &
  \includegraphics[width=40mm]{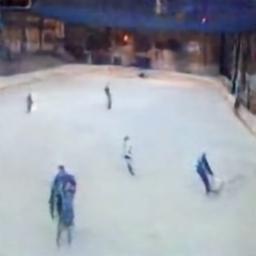} \\
  \includegraphics[width=40mm]{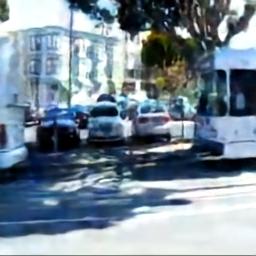} &
  \includegraphics[width=40mm]{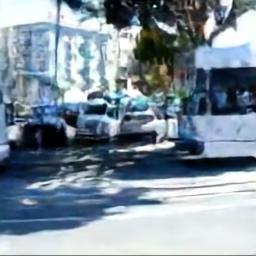} &
  \includegraphics[width=40mm]{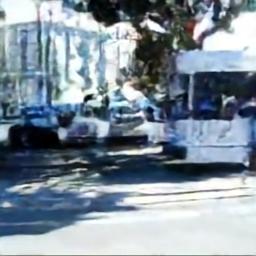} \\
  Final Conditioning Frame & 9th Frame & 16th Frame \\
  \end{tabular}
  \caption{Selected prediction results. The first 4 frames are given as conditioning. We predict the next 12, two of which (9th and 16th) we show on the right. All frames shown here have been compressed by VQ-VAE. Videos licensed under CC-BY. Attribution for videos in this paper can be found in section~\ref{Attributions}. Best seen in video on our website \href{https://sites.google.com/view/predicting-video-with-vqvae/home}{here}.}
\label{results2}
\end{figure}

In this section, we evaluate our model quantitively and qualitatively on the Kinetics-600 dataset \cite{Kinetics600}. This dataset of videos is very large and highly diverse, consisting of hundreds of thousands of videos selected from YouTube across 600 actions. While most previous work has focused on more constrained datasets, only a few \cite{clark2020adversarial, luc2020, WeissenbornTU20} have attempted to scale to larger size and complexity. We train our top and bottom models for around 1000000 iterations with a total batch sizes of 512 and 32 respectively. Our VQ-VAE model was trained on a batch size of 16 for 1000000 iterations. 

\subsection{Qualitative Evaluation}
While the Kinetics-600 dataset is publicly available for use, the individual videos in the dataset may not be licensed for display in an academic paper. Therefore, in this paper, we apply our model trained on Kinetics-600 to videos licensed under Creative Commons from the YFCC100m dataset \cite{yfc100m}. In figures \ref{results1} and \ref{results2} we show some selected predictions. We find that our approach is able to model camera perspective, parallax, inpainting, deformable human motions, and even aspects of crowd motion across a variety of different visual contexts. 

\subsection{Quantitative Evaluation}
Quantitative evaluation of generative models of images, especially across different classes of models, is an open problem \cite{TheisOB15}. It is even less explored in the realm of video prediction. One proposed metric used on larger datasets such as Kinetics-600 is the Fr\'echet Video Distance \cite{Unterthiner18}. As no previous approach has attempted $256\times256$ resolution, we downscale our videos to $64\times64$ for a proper comparison against prior work. We also compute FVD on the full resolution as a baseline for future work. We use 2-fold cross-validation over 39000 samples to compute FVD. We show our results in Table \ref{quant_results}. We find performance exceeds \cite{clark2020adversarial} but not necessarily \cite{luc2020}. We also find that comparing the VQ-VAE samples to the reconstructions, not the original videos, leads to an even better score (shown by FVD*). This result is similar to the results on images for VQ-VAE \cite{RazaviOV19}. As GAN-based approaches are explicitly trained on classifier (discriminator) based losses, FVD---a metric based on a neural-network classifier---may favor GAN-based approaches versus log-likelihood based models even if the quality of the samples are comparable. Given the possible flaws in this metric, we also conduct a human evaluation similar to \cite{VondrickPT16}. We had 15 participants compare up to 30 side-by-side videos generated from our approach and that of \cite{luc2020}. Each video had at least 13 judgements. For each comparison, both models used the exact set of conditioning frames and had a resolution of 64x64. Participants could choose a preference for either video, or they could choose indifference---meaning the difference in quality between two videos is too close to perceive. We show our results in Table \ref{quant_results2}. Out of a total of 405 judgements, participants preferred ours 65.7\% of the time, \cite{luc2020} 12.8\%, and 21.5\% were judged to be too close in quality. Even though \cite{luc2020} has a much lower FVD score, we find that our participants had stronger preference for samples generated from our model. 

\section{Conclusion}
In this paper we have explored the application of VQ-VAE towards the task of video prediction. With this learned compressed space, we can utilize powerful autoregressive models to generate possible future events in video at higher resolutions. We show that we are also able to achieve a level of performance comparable to contemporary GANs. 
\section*{Broader Impact}
Video prediction has the potential for broad, beneficial impact in a variety of situations. This includes superior video quality and speed via interpolation and compression, smoother human-computer interaction via implicit action forecasting, general improvements to model-based reinforcement learning, and cases in robotics where a model of the environment is needed. However, we are aware that this technology can be abused. Video generation could potentially be used to fabricate information or falsely portray individuals to deceptive or malicious social ends. 

There is already ongoing successful work in combating malicious applications of image and video generation---specifically the automatic detection of computer generated media \cite{afchar2018, sabir2019, xinsheng2019}. These types of detection efforts are particularly effective on the approach set out in this paper. By compressing videos through a bottleneck and removing high frequency details, our general videos are easily identifiable by an image classifier, reducing the risk of malicious use. As research in generative models progresses, it is essential that the field of visual forensics \cite{HuhLOE18}---progresses in tandem.

Finally, as \cite{clark2020adversarial} notes, showing video samples directly from the Kinetics dataset raises privacy concerns; the Kinetics dataset consists of videos uploaded from YouTube. Direct release of frames from this dataset raises questions of ethics, and generative models trained on this dataset could possibly memorize videos shown in the dataset. We address this issue and set a new precedent for future research by restricting qualitative results to videos licensed explicitly under Creative Commons by their authors. 

\section{Appendix}

\subsection{Architectures and Hyperparameters}
\begin{table}[h!]
  \caption{VQ-VAE Architecture Details}
  \centering
  \begin{tabular}{cc}
    \toprule
    Parameter     & Value \\
    \midrule
    Input Size & 256$\times$256$\times$16$\times$3 \\
    Latent layers & 32$\times$32$\times$4, 64$\times$64$\times$8 \\
$\beta$ (commitment loss coefficient) & 0.25 \\
Batch size & 16 \\
Hidden units & 128 \\
Residual units & 32 \\
Layers & 2 \\
Codebook size & 512 \\
Codebook dimension & 64 \\
First Stage Encoder twh-Conv Filter Size & 4 8 8 \\
First Stage Encoder twh-Conv Filter Stride & 2 4 4 \\
Second Stage Encoder twh-Conv Filter Size & 4 4 4 \\
Second Stage Encoder twh-Conv Filter Stride & 2 2 2 \\
Upsampling twh-Conv Filter Size & 4 8 8 \\
Upsampling twh-Conv Filter Stride & 2 4 4 \\
Training steps & 1000000 \\
    \bottomrule
  \end{tabular}
  \label{arch:1}
\end{table}

\begin{table}[t!]
  \caption{PixelCNN Prior Details}
  \centering
  \begin{tabular}{ccc}
    \toprule
    Parameter     & Top Prior & Bottom Prior \\
    \midrule
Input size & 32$\times$32$\times$3 & 64$\times$64 \\
Batch size & 512 & 32 \\
Hidden units & 512 & 512 \\
Residual units & 1024 & 1024 \\
Layers & 40 & 20 \\
Attention layers & 8 & 4 \\
Attention heads & 8 & 8 \\
Conv Filter size & 3 & 5 \\
Dropout & 0.5 & 0.0 \\
Training steps & 1016000 & 950000 \\ 
\bottomrule
  \end{tabular}
  \label{arch:2}
\end{table}

\begin{table}[t]
  \centering
  \begin{tabular}{cc}
    \toprule
    Parameter     & Values \\
    \midrule
Input size & 32$\times$32 (upsampled to 64$\times$64$\times$2), 64$\times$64$\times$2 \\
Hidden units & 512  \\
Residual units & 128 \\
Layers & 4 \\
\bottomrule
\\
  \end{tabular}
  \caption{Top Prior Conditioning Stack. The conditioning frames 256$\times$256$\times$4$\times$3 are compressed using our VQ-VAE into a 32$\times$32 and 64$\times$64$\times$2 space. These two layers are then concatenated, downsampled back down to 32$\times$32$\times$256 by a convolutional layer, tiled to 32$\times$32$\times$3$\times$256, and finally fed through another convolutional layer to 32$\times$32$\times$3$\times$512 before being fed through four residual blocks.}
\end{table}

\begin{table}[t!]
  \centering
  \begin{tabular}{cc}
    \toprule
    Parameter     & Values \\
    \midrule
Input size & 32$\times$32$\times$4, 64$\times$64$\times$4 \\
Hidden units & 1024  \\
Residual Blocks & 20 \\
\bottomrule
\\
  \end{tabular}
  \caption{Bottom Prior Conditioning Stack. Let $n$ be the timestep to be modeled by the bottom prior. The 32$\times$32$\times$4 top layer is upsampled to 64$\times$64$\times$8$\times$1024 through a series of three convolutional layers with kernel sizes (4, 3, 3)$\rightarrow$(3,4,4)$\rightarrow$(3, 3, 3) and strides (2, 1, 1)$\rightarrow$(1, 2, 2)	$\rightarrow$(1, 1, 1) respectively. From this output, the $n$th slice is chosen as input. For the bottom layer, an input the past $n-4, n-3, ... n-1$ timesteps are fed into a series of 4 convolutional layers at size (4, 3, 3) and stride (2, 1, 1) each. This downsamples to a layer of size 64$\times$64. This is concatenated with the output from the top layer and fed into a conditioning stack identical to the one described in \cite{RazaviOV19}. }
\end{table}

\clearpage

\subsection{Attributions} \label{Attributions}

All videos shown in this paper are licensed under Creative Commons BY 2.0 which can be accessed \href{https://creativecommons.org/licenses/by/2.0/}{here}. All videos have been cropped and modified by our algorithm.

Figure~\ref{teaser}: \\
“ Ocean Beach Surfing” by dakine kane.
Accessible \href{https://www.flickr.com/photos/61112791@N00/8361264621}{here}

Figure~\ref{reconst}: \\
“ Ocean Beach Surfing” by dakine kane.
Accessible \href{https://www.flickr.com/photos/61112791@N00/8361264621}{here}

Figure~\ref{VQVAE}: \\
“ Ocean Beach Surfing” by dakine kane.
Accessible \href{https://www.flickr.com/photos/61112791@N00/8361264621}{here}

Figure~\ref{results1}: \\
“Cutting Shrimp” by jencu.
Accessible \href{https://www.flickr.com/photos/10581108@N00/3767578041}{here} \\
“New York City December 2010” by Duncan Currier.
Accessible \href{https://www.flickr.com/photos/26844488@N07/5297091231}{here} \\
“How to wrap beef currie pies” by Joy.
Accessible \href{https://www.flickr.com/photos/33993074@N00/3120549283}{here} \\
“dave skiing 2” by Leigh Blackall.
Accessible \href{https://www.flickr.com/photos/97283472@N00/3072947082}{here} \\
“Ashokan Farewell played as a trio” by Bryn Pinzaguer.
Accessible \href{https://www.flickr.com/photos/12394349@N06/4181870078}{here} \\

Figure~\ref{results2} \\
“Dan skiing.” by Deana Hunter:
Accessible \href{https://www.flickr.com/photos/87013411@N00/3893217499}{here} \\
“Every Saturday in Canada is Hockey night - watching this makes me want to play hockey
again!” by Roland Tanglao.
Accessible \href{https://www.flickr.com/photos/35034347371@N01/4448745439}{here} \\
“The Cable Car” by Aaron Tait.
Accessible \href{https://www.flickr.com/photos/96272984@N00/2442344223}{here} \\

\small
\printbibliography

\end{document}